\documentclass{article} 
\usepackage{arxiv,times}

\usepackage{amsmath,amsfonts,bm}

\def\eqref#1{equation~\ref{#1}}

\def\1{\bm{1}}

\DeclareMathAlphabet{\mathsfit}{\encodingdefault}{\sfdefault}{m}{sl}
\SetMathAlphabet{\mathsfit}{bold}{\encodingdefault}{\sfdefault}{bx}{n}

\usepackage{hyperref}
\usepackage{url}
\usepackage{xspace}
\usepackage{times}
\usepackage{latexsym}
\usepackage{graphicx} 
\usepackage[T1]{fontenc}

\usepackage[utf8]{inputenc}
\usepackage{booktabs} 
\usepackage{multirow}

\usepackage{microtype}

\usepackage{inconsolata}
\usepackage{tikz}
\usepackage{pgf-pie}
\usepackage{xcolor}
\usepackage{float}
\usepackage{ulem}
\usepackage{makecell}
\usepackage{multirow}
\usetikzlibrary{shadows}

\title{At Which Training Stage Does Code Data Help LLMs Reasoning?}

\author{Yingwei Ma~$^*$\\
National University of Defense Technology\\
Peng Cheng Laboratory\\
\texttt{yingwei.ywma@gmail.com} \\
\And
Yue Liu \thanks{co-first author.}\\
National University of Defense Technology\\
\texttt{yueliu19990731@163.com} \\
\And
Yue Yu~\thanks{corresponding author.}\\
National University of Defense Technology \\
Peng Cheng Laboratory \\
\texttt{yuyue@nudt.edu.cn} \\
\And
Yuanliang Zhang \\
National University of Defense Technology \\
\texttt{zhangyuanliang13@nudt.edu.cn} \\
\And
Yu Jiang \\
Tsinghua University \\
\texttt{jiangyu198964@126.com} \\
\And
Changjian Wang \\
National University of Defense Technology \\
\texttt{changjianwang@nudt.edu.cn} \\
\And
Shanshan Li~$^\dag$ \\
National University of Defense Technology \\
\texttt{shanshanli@nudt.edu.cn} \\
}

\newcommand{\ie}{{\textit{i.e.}},\xspace}
\newcommand{\eg}{{\textit{e.g.}},\xspace}

\begin{document}

\maketitle

\begin{abstract}
Large Language Models (LLMs) have exhibited remarkable reasoning capabilities and become the foundation of language technologies. Inspired by the great success of code data in training LLMs, we naturally wonder at which training stage introducing code data can really help LLMs reasoning. To this end, this paper systematically explores the impact of code data on LLMs at different stages. Concretely, we introduce the code data at the pre-training stage, instruction-tuning stage, and both of them, respectively. Then, the reasoning capability of LLMs is comprehensively and fairly evaluated via six reasoning tasks in five domains. We critically analyze the experimental results and provide conclusions with insights. First, pre-training LLMs with the mixture of code and text can significantly enhance LLMs' general reasoning capability almost without negative transfer on other tasks. Besides, at the instruction-tuning stage, code data endows LLMs the task-specific reasoning capability. Moreover, the dynamic mixing strategy of code and text data assists LLMs to learn reasoning capability step-by-step during training. These insights deepen the understanding of LLMs regarding reasoning ability for their application, such as scientific question answering, legal support, etc. The source code and model parameters are released at the link:~\url{https://github.com/yingweima2022/CodeLLM}.
\end{abstract}

\section{Introduction}

Recently, Large Language Models (LLMs) have achieved impressive generalization performance across various tasks. Significantly, OpenAI developed ChatGPT~\cite{chatgpt}, Google designed PaLM~\cite{chowdhery2022palm}, Baidu built ERNIE Bot~\cite{wenxin}, and Alibaba presented Tongyi Qianwen~\cite{tongyi}. However, these industrial products are regrettably not open-source for commercial reasons. Thanks to the surging open-source projects of LLMs such as LLaMA~\citep{touvron2023llama}, Alpaca~\citep{alpaca}, and GLM~\citep{DBLP:conf/acl/DuQLDQY022}, the academic research and industrial products of LLMs mark new milestones.

Two of the key factors to the great success of LLMs are 1) training data and 2) training strategies. First, for the training data, researchers aim to endow LLMs with language capabilities and general knowledge via training models on large-scale data from various domains. For example, LLaMA was trained with 1.4 trillion tokens consisting of texts (CommonCrawl, C4) and codes (GitHub). These large-scale data with diversity help the model to achieve competitive performance on multiple tasks. Second, the common pipeline goes through two stages for the training strategies: pre-training and instruction-tuning. The pre-training is conducted in a self-supervised manner on the massive unlabeled data, while instruction-tuning aims to fine-tune models with human-annotated prompts and feedback~\citep{ouyang2022training}. Benefiting from the data and training strategies, LLMs gain remarkable skills, such as translation, conversation, examination, legal support, etc. These skills are all based on one of the most important capabilities, i.e., reasoning capability. So, how can LLMs gain such strong reasoning capability?

We analyze the reasons from two aspects: training data and strategies. First, from the training data aspect, compared with the common textual data, code data is more logical and less ambiguous (refer to case studies in Appendix \ref{casestudy}). Also, from the experiments, researchers~\citep{liang2022holistic, fu2022gptroadmap} verified that models trained on code data have strong reasoning capability. Therefore, code data is essential for model reasoning. Second, for the training strategies, both pre-training and fine-tuning are crucial to the model's performance. Pre-training feeds general knowledge to models while fine-tuning feeds domain-specific ability to models. To further explore the deep-in reasons for the strong reasoning capability of LLMs, this paper aims to answer an important question: at which stage does code data help LLMs reasoning?

To this end, we conduct comprehensive and fair experiments and provide analyses and conclusions with insights. First, we pre-train LLMs with pure text data and mixture data of code and text, respectively. Subsequently, at the instruction-tuning stage, LLMs are fine-tuned with the pure text data and mixture data of code and text, respectively. After training, to comprehensively measure the model reasoning capability, we evaluate LLMs on six tasks in five domains, including logical reasoning, code reasoning, legal reasoning, scientific reasoning, and analogical reasoning. Based on extensive experimental results and analyses, we provide three insights. 1) Pre-training LLMs with the mixture of code and text can significantly enhance LLMs' general reasoning capability almost without negative transfer on other tasks. 2) At the instruction-tuning stage, code data endows LLMs the task-specific reasoning capability. 3) The dynamic mixing strategy of code and text data assists LLMs to learn reasoning capability step-by-step during training. These findings deepen the understanding of LLMs regarding reasoning ability for their applications, such as scientific question answering, legal support, etc. The main contributions of this work are summarized as follows.

\begin{itemize}
    \item Research question: this paper raises and aims to answer one essential concern, i.e., at which training stage can codes data help LLMs reasoning. 
    
    \item Analyses and insights: we conduct extensive experiments and provide critical analyses and insights, which deepen the understanding of LLMs regarding reasoning capability.

    \item Open-source resource\footnote{\url{https://anonymous.4open.science/r/CodeLLM-FD25/}}: we release the model implementation and the trained model parameters, which contribute to the further research in the LLMs community.
    
\end{itemize}

\subsection{Training Data \& Training Strategies}

Three key factors to the great success of LLMs are training data, training strategies, and model designs. In this section, we introduce our training data and training strategies. The next section details the model designs. 



\begin{figure}[h!]
    \centering
    \includegraphics[scale=0.32]{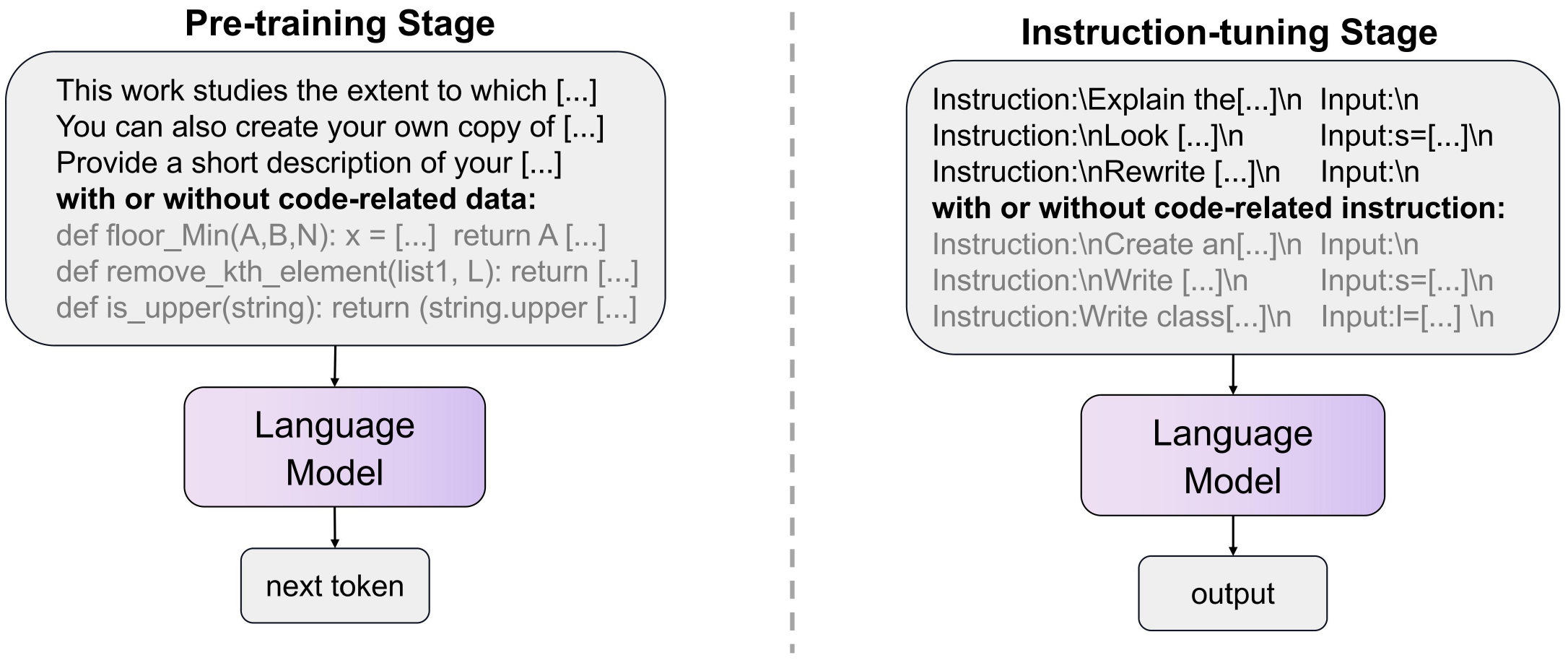}
    \caption{Demonstration of the pre-training and tuning phase.}
    \label{fig:fig1}
\end{figure}

We study two training phases of LLMs, i.e., pre-training stage and instruction-tuning stage, on two different datasets including one plain text data and one text-code-mixed data. Figure \ref{fig:fig1} demonstrates the process of each stage. Specifically, we use the open-sourced PanGu2.6B and PanGu13B of the PanGu-$\alpha$ team~\cite{zeng2021pangu} as baseline models for text models (trained on 100GB text data and larger text data, respectively), and train CodePanGu2.6B from scratch on the mixed code data for comparison. We will introduce detailed data settings in later chapters.

\subsection{Pre-Training Corpus}\label{pretrainsetup}

\begin{figure}[h]
\centering
\small
\begin{minipage}{1.0\linewidth}
\centerline{\includegraphics[scale=0.30]{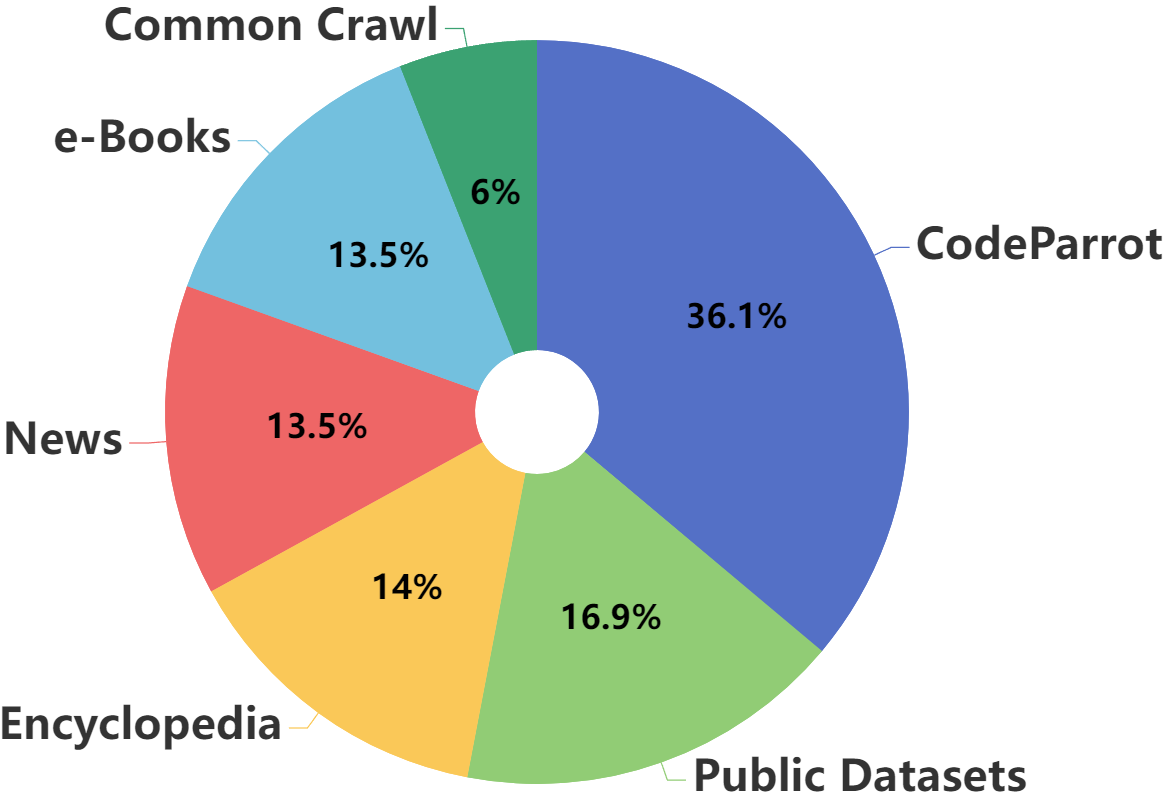}}
\end{minipage}
\caption{Distribution of the $\sim$42B tokens in pre-training data.}
\label{contrastive}
\end{figure}

The pre-training corpus consists of two parts. To ensure a fair comparison with PanGu2.6B, we collected a large amount of original data from public datasets such as BaiDuQA, CAIL2018, Sogou-CA, and network data sets such as Common Crawl, encyclopedias, news, and e-books according to the PanGu-$\alpha$ team~\citep{zeng2021pangu}. Then we use rule-based data cleaning and model-based data filtering methods to filter to ensure high quality. Finally, we obtain 100GB of text data with the same scale and source as PanGu2.6B by sampling each data source using different ratios. Please refer to Appendix \ref{pangudata} for a detailed data processing process. To verify the influence of code data on the reasoning capability of the model in the pre-training stage, we used the CodeParrot~\citep{codeparrot} dataset as the second supplementary part. CodeParrot is a public Python dataset from BigQuery, comprising approximately 50GB of code and 5,361,373 files. Figure \ref{contrastive} shows the composition of the $\sim$42B tokens in pre-training data.

\subsection{Instruction-Tuning Corpus} \label{instructionsetup}

We collect and construct 500K instruction tuning data to verify the effect of adding code instructions in the instruction tuning stage and convert them into a unified instruction format. The instruction tuning corpus is divided into two parts. The first part is from the natural language open source instruction dataset, Alpaca-GPT-4~\citep{peng2023instruction} and PromptCLUE~\citep{pCLUE}. Alpaca-GPT-4 is generated by GPT-4, including 52K Chinese and English instruction tuning data. PromptCLUE unifies the differences between different NLP tasks (\eg reading comprehension, question answering) and converts the original task training set into a unified text-to-text data form, from which we randomly sample 200K data for instruction tuning.

\begin{figure}
    \centering
    \includegraphics[scale=0.46]{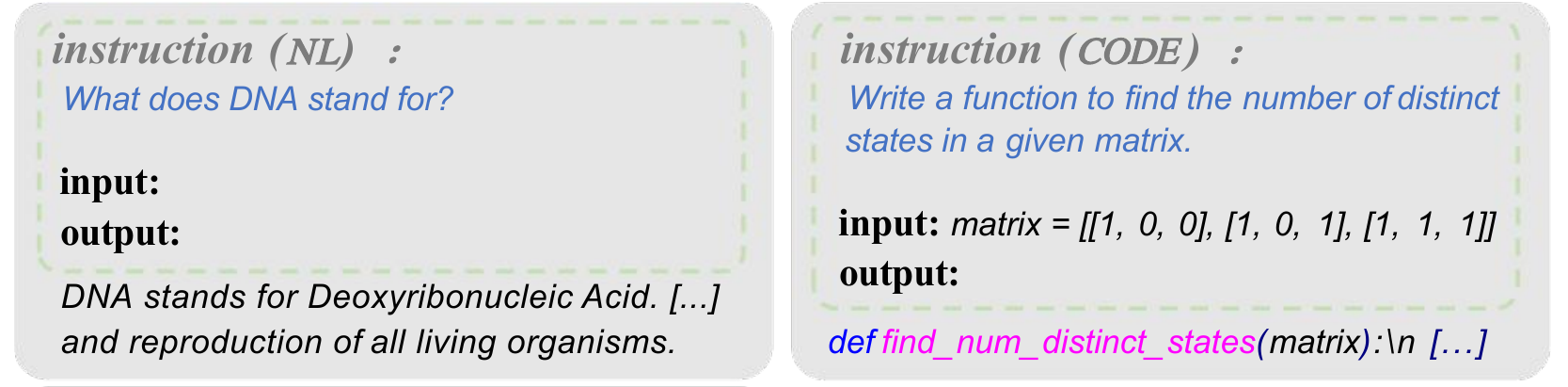}
    \caption{Example of the instruction tuning data format. NL denotes natural language. }
    \label{fig:method1}
\end{figure}

The second part comes from the open-source data CodeAlpaca~\citep{codealpaca} and our build dataset, with 150K instructions. The CodeAlpaca data contains 20K instruction tuning data generated according to the self-instruct technology, which can be used for instruction tuning of the code generation model. In order to supplement the code-related instruction tuning data, we use the CosQA~\citep{huang2021cosqa} training set and the MBPP~\citep{austin2021program} training set to unify the task format in the way of PromptCLUE and expand the CodeAlpaca data. Figure \ref{fig:method1} is an example of the format of instruction tuning data.

\section{Model}

We conduct experiments on large-scale autoregressive language models by adopting the GPT paradigm\citep{brown2020language}. It iteratively takes all tokens in the corpus as input, predicts the next token, and compares it to the ground truth. Assuming that a sequence $\mathcal{X} = \{x_{1}, x_{2}, ..., x_{n}\}$ is composed of $n$ tokens, the training objectiv
e can be formulated as maximization of the log-likelihood:
\begin{equation}
\mathcal{L} = \sum_{i=1}^{n}\log p(x_{i}|x_{1}, x_{2}, ...,x_{i-1};\Theta)
\end{equation}

where $p(x_{i}|x_{1}, x_{2},...,x_{i-1};\Theta)$ is the probability of observing the $i$-th token $x_{i}$ given the previous context $x_1, x_2, ..., x_{i-1}$, and $\Theta$ denotes the model parameters.

\subsection{Model Architecture}

Similar to recent pre-trained models such as GPT-3~\cite{brown2020language}, LLaMA~\citep{touvron2023llama}, CodeGeeX~\citep{zheng2023codegeex}, and PANGU-$\alpha$~\citep{zeng2021pangu}, we follow a generative pre-training (GPT) architecture for autoregressive language modeling. At the same time, to make a fair comparison with the baseline of PanGu2.6B, we retain the setting of the 32-layer transformer decoder. The original GPT model uses a pooler function to obtain the final output. Follow CodeGeeX~\citep{zheng2023codegeex} and PANGU-$\alpha$~\citep{zeng2021pangu}, we use an additional query layer on top of the stacked Transformer layers to explicitly induce the expected output with attention to obtain the final embedding.

\subsection{Tokenization}
For the text-only model, we use the open-source vocabulary of the PanGu2.6B model released by PanGu-$\alpha$ team~\citep{zeng2021pangu}, and the size of the vocabulary is 40,000. For the model training with mixed code, considering that there may be variables, functions, and class names in the code that are often meaningful words, we use the ChatGLM~\citep{du2022glm} vocabulary open-sourced by the THUGLM team to encode text and the code. The vocabulary size is 130,044. In addition, ChatGLM encodes multiple spaces as extra tokens to improve encoding efficiency. Specifically, L spaces are represented by <|extratoken\_X|>, where X=8+L. Both vocabularies are BPE-based tokenizers, which use fixed-size vocabularies to handle variable-length characters in open-vocabulary problems.

\section{Experiments}
\subsection{Task Description}

To measure the reasoning ability of the models, we evaluate it on six tasks in realistic reason-centric scenarios, including general reasoning scenarios such as logical reasoning, legal reasoning, scientific reasoning, and analogical reasoning, and code-related scenarios such as code generation. These reasoning-intensive tasks elucidate the reasoning capabilities of the model through the model's performance in these scenarios. When publicly available, we evaluate the models with the test sets for each task. Otherwise, we use the development sets instead. We describe each task as follows.

{\bfseries Logical Reasoning.} Logic is the study of reasoning and argumentation, which focuses on the rules of logic and methods of reasoning in the thinking process. We use the {\bfseries logic} subject in the {\bfseries C-Eval} dataset~\citep{huang2023ceval} to determine whether the model can understand and apply logical rules to make reasonable reasoning.

{\bfseries Legal Reasoning.} For legal reasoning, we use {\bfseries JEC-QA}~\citep{zhong2019jec},  the largest question answering dataset in the legal domain, collected from the National Judicial Examination of China. The examination is a comprehensive evaluation of the professional skills of legal practitioners. Multiple reasoning skills are required to retrieve relevant material and answer legal questions.

{\bfseries Scientific Reasoning.} We use the {\bfseries ScienceQA} dataset~\citep{lu2022learn} to evaluate the scientific reasoning ability of the model. The scientific question answering task can diagnose whether the artificial intelligence model has multi-step reasoning ability and interpretability. To answer scientific questions from ScienceQA, a model not only needs to understand multimodal content but also needs to extract external knowledge to arrive at the correct answer.

{\bfseries Analogical Reasoning.} We use the {\bfseries E-KAR} dataset~\cite{chen-etal-2022-ekar} to evaluate the model's analogical reasoning ability. It comes from the Civil Service Examination, a comprehensive test of the candidate's critical thinking and problem-solving ability. To solve the analogy reasoning problem, candidates need to understand the relationship among the options, which requires specific reasoning ability and background knowledge, especially common sense and facts, and knowing why a fact is denied.

{\bfseries Code Reasoning.} We use {\bfseries CosQA}~\citep{huang2021cosqa} to test the model performance on the code question-answering task. The dataset includes 604 natural language-code question-answer pairs. Furthermore, we use the {\bfseries MBPP} dataset~\citep{austin2021program} to test the model code generation ability, containing 427 Python coding questions.

\subsection{Evaluation Details} \label{evaluationdetails}

\begin{table*}
\renewcommand{\arraystretch}{1.3}
\centering
\scalebox{0.90}{
\begin{tabular}{lcc}
\hline
\textbf{Task Type} & \textbf{Dataset} &\textbf{Input \& Prompt}\\
\hline
Logical & Logic & The answer: \underline{\$choice}, can answer the following questions: \underline{\$problem} \\
Legal & JEC-QA & The answer: \underline{\$choice}, can answer the following questions: \underline{\$problem} \\
Scientific & ScienceQA & \underline{\$lecture}\textbackslash n anwser: \underline{\$choice} can answer the following question: \underline{\$question} \\
Analogical & E-KAR & The reasoning relationship: \underline{\$r1}, the analogy reasoning relationship: \underline{\$r2} \\
Code & CosQA & \underline{\$question}? Answered code is correct or wrong: \$code \\
Code & MBPP & \underline{\$question}\textbackslash n Code:\textbackslash n \\ \hline 
\end{tabular}}
\caption{The input \& prompt template for each task. \underline{\$       } is the input and other words are prompt. }
\label{tab:prompt}
\end{table*}

\begin{table*}
\renewcommand{\arraystretch}{1.3}
\centering
\scalebox{1}{
\begin{tabular}{lccccc}
\hline
\textbf{Dataset} & \textbf{Task} & \textbf{Metric} &\textbf{NL (2.6B)} &\textbf{NL (13B)} &\textbf{CODE (2.6B)}\\
\hline
Logic$^*$ & Logical Reasoning & ACC & 36.36 & \textbf{45.45} & 40.90 \\
JEC-QA$^*$ & Legal QA & ACC & 27.00 &  27.00 & \textbf{28.70}\\
ScienceQA$^*$ & Scientific QA & ACC & 45.93 & 45.18 & \textbf{46.06}\\
E-KAR$^*$ &Analogical Reasoning & ACC & 32.24 & 35.52 & \textbf{36.12} \\
CosQA$^\dag$ & Code QA & ACC & 47.01 & 46.85 & \textbf{50.50} \\
MBPP$^\dag$ &Code Generation & BLEU & 0.52 & 1.34 & \textbf{5.06}  \\ \hline
\end{tabular}}
\caption{Results on pre-training stage. Bold values indicate the best performance. $^*$ denote the general reasoning task, and $^\dag$ denote the code-related reasoning task.}
\label{tab:res_pretrain}
\end{table*}




In evaluation, these tasks are usually divided into two parts, understanding task and generation task. For the understanding task, we follow PanGu2.6B~\citep{zeng2021pangu} and CPM~\cite{zhang2021cpm}, decomposing the task into a perplexity comparison task. We construct a prompt template for each evaluation task and populate the template with instances as input to the model. Table \ref{tab:prompt} describes the templates for each task.


We adopt a perplexity-based approach to solve classification tasks. For each <text, label> pair, input will be automatically generated according to the predesigned prompt in Table \ref{tab:prompt}. The sequences generated by the prompt will be fed into the model, and a perplexity value will be calculated. The label corresponding to the minimum perplexity value will be regarded as the predicted label for this passage. For the generative task, we leverage the properties of autoregressive language models to generate corresponding answers directly from a given input naturally.

\subsection{Results}

\subsubsection{Pre-training Stage}

To illustrate the impact of code data in the pre-training phase on the reasoning capabilities of large language models, we compared the performance of the three models in real reasoning-intensive scenarios. Among them, the NL (2.6B) and NL (13B) (\ie PanGu2.6B and PanGu13B) models~\citep{zeng2021pangu} are trained on natural language datasets, and the CODE (2.6B) (\ie CodePangu2.6B) model is trained on mixed data (the dataset mentioned in Chapter \ref{pretrainsetup}). The models are evaluated in zero-shot manner on downstream tasks. Specifically, we report accuracy on for Logic, JEC-QA, ScienceQA, E-KAR, and CosQA tasks and BLEU score for MBPP task. Table \ref{tab:res_pretrain} depicts the results of these tasks. Consistently over these tasks, we have two observations as follows. 


\begin{itemize}
    \item After adding code training, LLM performs better on most reasoning-related tasks, even though most of these tasks are not related to code. This shows that adding code data in the pre-training stage can not only improve the coding-related ability but also improve the general language reasoning ability of the model to a certain extent.
    
    \item Even with a larger scale model, \ie NL (13B), it is still not as effective as CODE (2.6B) in these reasoning scenarios. This is similar to the results of HELM~\cite{liang2022holistic}, which suggest that if (a) the computational budget is constrained and (b) the resulting model is applied in the code reasoning domain, adding code data in the pre-training phase may be more effective than increasing the model parameter size.
\end{itemize}

In summary, we find that simply adding code data during the pre-training phase can effectively improve the model's general reasoning ability, which might indicate that mixing more code data for training may produce a competitive model to solve tasks that require complex reasoning to complete. This provides a promising prospect for subsequent LLM development.

\subsubsection{Instruction-tuning Stage}

\begin{table*}
\centering
\renewcommand{\arraystretch}{1.3}
\scalebox{1.0}{
\begin{tabular}{lccccc}
\hline
\textbf{Dataset} &  \textbf{NN (2.6B)} & \textbf{NC (2.6B)}& \textbf{NN (13B)}& \textbf{NC (13B)} & \textbf{CC (2.6B)}\\
\hline
Logic$^*$ & 36.36  & {40.90} & {40.90} & {40.90} & \textbf{40.90} \\
JEC-QA$^*$ & 25.20 & 26.10 & 24.50 & 26.40 & \textbf{27.10}\\
ScienceQA$^*$   & \textbf{44.45}  & 43.44  & 42.94 & 43.41 & 41.90\\
E-KAR$^*$   & \textbf{30.45}  & 28.66 & 26.27 & 27.46 & 27.20\\
CosQA$^\dag$ & 45.20 & 48.18  & 47.52  & 51.99 & \textbf{52.48} \\
MBPP$^\dag$  & 0.00 & 5.61 & 0.00 & 1.88 & \textbf{24.88} \\ \hline
\end{tabular}}
\caption{Results on instruction-tuning stage. Bold values indicate the best performance. $^*$ denote the general reasoning task, and $^\dag$ denote the code-related reasoning task. }
\label{tab:instruction}
\end{table*}


ChatGPT~\cite{chatgpt} and GPT-4~\cite{gpt4} successfully use instruction tuning to enable LLMs to follow natural language instructions and complete real-world tasks; this improvement has become standard in open-source LLMs. This is implemented by fine-tuning the model on a wide range of tasks using human-annotated instructions and feedback, by supervised fine-tuning via manually or automatically generated instructions using public benchmarks and datasets, or learning from instruction-following data by developing from state-of-the-art instruction-tuned teacher LLMs.

To illustrate the impact of code data on the LLMs reasoning ability in the instruction tuning stage, we use the instruction tuning datasets that contain codes and the instruction tuning datasets without codes introduced in Chapter \ref{instructionsetup} to fine-tune the PanGu2.6B and PanGu13B models~\citep{zeng2021pangu} and evaluate their performance in reasoning-intensive scenarios. In addition, we also fine-tune the CodePanGu2.6B model using the instruction tuning dataset containing codes to observe the effect of using code data in both pre-training and instruction tuning stages. Table \ref{tab:instruction} shows the results of these tasks. Among them, NN and NC represent the fine-tuned PanGu model using only text instructions and instructions containing codes, respectively, and CC represents the fine-tuning model of CodePanGu2.6B using instructions containing codes. Consistently over these tasks, we observe the following:

\begin{itemize}
    \item After fine-tuning with mixed code instruction data, LLM shows different trends in multiple reasoning tasks. This indicates that introducing code data in the instruction tuning phase may be less effective than in the pre-training phase. Therefore, it is best to add code data in the pre-training stage to improve the model performance in general reasoning tasks.
    
    \item We find that training with code data in both stages can significantly improve code-related tasks~(CosQA and MBPP), especially code generation tasks. This may be because the code instruction data activates the code reasoning ability of the language model, which suggests that if the LLM needs to complete complex code tasks, the code reasoning ability can be improved by effectively following code instructions and generating compliant content.
    
    \item Compared with the pre-training stage, the performance of instruction-tuned LLMs on some tasks is degraded, similar to the TÜLU~\citep{wang2023far} results. This may be because the instruction tuning data usually covers a wide range of domains and dialogue content, causing the model to tend to answer questions more comprehensively, resulting in a decline in reasoning ability. We propose that if specific reasoning capabilities are required, they can be augmented by adding domain-specific instructions during the tuning phase.
\end{itemize}

In summary, we find that adding code data in the instruction tuning stage is not as effective as the pre-training stage in improving the general reasoning ability of the model. However, we find that code instructions made the model follow natural language instructions and generate correct code, improving the model's code reasoning ability. This also suggests that tuning with relevant data may be helpful when solving specific reasoning tasks.

\subsubsection{Chain-of-Thought Ability}

Compared with the standard prompt technology, Chain-of-Thought (CoT)~\citep{wei2022chain} transforms tasks into a continuous chain generation process. This technology enhances the model ability in complex reasoning tasks by providing a language model with a series of related reasoning steps. To evaluate the potential of the model in utilizing chains of thought in solving complex problems, we conduct experiments on two pre-trained models, NL (2.6B), \ie PanGu2.6B and CODE (2.6B), \ie CodePanGu2.6B on ScienceQA(CoT)~\citep{lu2022learn} and E-KAR(CoT)~\citep{chen-etal-2022-ekar} datasets. We incorporate CoT information as a part of the model input with the question and context information. In this way, the model can directly use the reasoning process of the thinking chain for answer generation. The experimental results are shown in Table \ref{tab:cot}.

The experimental results show that after the introduction of the Chain-of-Thought, the performance of all models in reasoning problems is significantly improved by making full use of the coherent reasoning process of CoT. The CoT information is used as part of the model input to help the model better understand the problem and generate answers according to the logic of the CoT. Among them, CODE (2.6B) achieved the best performance, indicating that CODE (2.6B) can better use CoT information for reasoning. This also suggests that pre-training with mixed-code data may result in a competitive model for tasks that require complex reasoning.

\begin{table}
\centering
\renewcommand{\arraystretch}{1.3}
\scalebox{1.0}{
\begin{tabular}{lccc}
\hline
\textbf{Model} & \textbf{Dataset} & \textbf{without CoT} & \textbf{with CoT} \\
\hline
NL (2.6B) & ScienceQA & 45.93  & 68.76 \\
CODE (2.6B) & ScienceQA & \textbf{46.06} & \textbf{70.30}  \\
NL (2.6B) & E-KAR & 32.24 & 69.55 \\
CODE (2.6B) & E-KAR & \textbf{36.12} & \textbf{72.84}  \\ \hline
\end{tabular}}
\caption{Results with Chain-of-Thought prompt. Bold values denote the best results.}
\label{tab:cot}
\end{table}

\subsubsection{Exploring Ways to Mix Code and Text Data}
\label{sec:mixway}

Previous experiments have demonstrated that training with mixed code data in the two stages of pre-training and instruction tuning can improve the general and specific reasoning capabilities of LLMs, respectively. Therefore, We naturally wonder how mixing these two types of data can better improve model reasoning ability, which has not been explored in previous studies. Therefore, we design comparative experiments in the instruction tuning stage to verify the impact of different data mixing strategies. The mixed strategy is shown in Table \ref{tab:mixdata}. One group is uniform sampling, that is, the proportion of text and code in each group of training data is roughly the same; the other two groups gradually increase or decrease the proportion of code to verify whether step-by-step learning will better activate the reasoning ability of LLMs. The experimental results are shown in Table \ref{tab:mixdatares}.

\begin{table}[h]
\centering
\renewcommand{\arraystretch}{1.3}
\scalebox{1.0}{
\begin{tabular}{lccc}
\hline
\textbf{Phase} & \textbf{{\makecell[l]{Uniform Sampling}}} &\textbf{{\makecell[l]{Stepwise Increase}}} &\textbf{{\makecell[l]{Stepwise Decrease}}} \\
\hline
1 & 5:3 & 7:3 & 5:5\\
2 &  5:3 & 7:3 & 6:4\\
3 & 5:3 & 6:4  & 7:3\\
4 & 5:3 & 5:5 & 7:3\\ \hline
\end{tabular}}
\caption{Mixture strategies on text data and code data with different ratios (text:code).}
\label{tab:mixdata}
\end{table}


\begin{table}[h]
\centering
\renewcommand{\arraystretch}{1.3}
\scalebox{1.0}{
\begin{tabular}{lccc}
\hline
\textbf{Dataset} &  \textbf{{\makecell[l]{Uniform Sampling}}} &\textbf{{\makecell[l]{Stepwise Increase}}} &\textbf{{\makecell[l]{Stepwise Decrease}}} \\
\hline
Logic$^*$  & 31.82 & 36.36 & \textbf{40.90}\\
JEC-QA$^*$ & \textbf{27.30} & 26.70 &  27.10\\
ScienceQA$^*$ & \textbf{43.76} & 43.19 & 41.90\\
E-KAR$^*$ & \textbf{28.66} & 28.36 & 27.20 \\
CosQA$^\dag$ & 51.65 & 50.66 &  \textbf{52.48}\\
MBPP$^\dag$ & 23.68 & 23.42 & \textbf{24.88} \\ \hline
\end{tabular}}
\caption{Result of different mixed strategies. Bold values indicate the best performance. $^*$ denote the general reasoning task, and $^\dag$ denote the code-related reasoning task. }
\label{tab:mixdatares}
\end{table}

The experiment found that the training strategy of using a higher code data ratio in the early stage and gradually reducing the code data ratio in the later stage achieved the best results in code question answering~(CosQA) and code generation~(MBPP) tasks, while ensuring the performance of the model in other reasoning tasks. This may be because, due to the strong logic of the code, using more code data in the early stage may help the model activate the code reasoning ability faster. Therefore, if LLMs are expected to have better specific reasoning ability, adopting a stepwise descent strategy can better activate the model potential. In addition, since experiments in the pre-training phase require a lot of resources, we leave the validation of this phase to later work.

\subsubsection{Other Tasks}

\begin{table}
\centering
\renewcommand{\arraystretch}{1.2}
\scalebox{1.0}{
\begin{tabular}{lccc}
\hline
\textbf{Dataset} & \textbf{Metrics} & \textbf{without code} & \textbf{with code} \\
\hline
$C^3$ & ACC & 54.14  & \textbf{54.30} \\
OCNLI & ACC  & \textbf{41.69} & 40.50  \\
CMNLI & ACC & \textbf{45.07} & 43.49 \\
\multirow{2}{*}{DuReader} & EM & \textbf{0.42} & 0.14  \\
& F1 & \textbf{15.29} & 8.73  \\  \hline
\end{tabular}}
\caption{Results of pre-training on other tasks. Bold values indicate the best performance.}
\label{tab:OtherTasks_pretrain}
\end{table}

\begin{table}
\centering
\renewcommand{\arraystretch}{1.2}
\scalebox{1.0}{
\begin{tabular}{lccc}
\hline
\textbf{Dataset} & \textbf{Metrics} & \textbf{without code} & \textbf{with code} \\
\hline
$C^3$ & ACC & \textbf{55.07}  & 54.47 \\
OCNLI & ACC  & 40.78 & \textbf{41.19}  \\
CMNLI & ACC & 44.82 & \textbf{45.49} \\
\multirow{2}{*}{DuReader} & EM & \textbf{12.07} & 8.05  \\
& F1 & \textbf{34.85} & 25.05  \\ \hline
\end{tabular}}
\caption{Results of instruction-tuning on other tasks. Bold values indicate the best performance.}
\label{tab:OtherTasks_finetune}
\end{table}

We extensively evaluates various reasoning tasks, including logical and code reasoning, highlighting the positive impact of code-related data. Additionally, we sought to ascertain whether code data would affect common-sense tasks. Therefore, to verify the impact of code data on other comprehension and generation tasks that are less demanding on reasoning, we conduct experiments on other tasks, including two NLI tasks (OCNLI~\citep{ocnli} and CMNLI~\citep{wang2018glue}), requiring the model to identify the relationship between two sentences, either entailment, neutral or contradiction; a free-form multiple-choice Chinese machine reading comprehension dataset ($C^3$)~\citep{sun2020investigating} consisting of documents (conversational or more formal mixed-type text) and their associated multiple-choice free-form questions; one reading comprehension task duReader~\citep{he2017dureader}, requiring the model to extract a text span from a given paragraph as the correct answer to the question. Refer to Appendix \ref{sec:appendixA} for prompt templates and evaluation metrics for different tasks.

Table \ref{tab:OtherTasks_pretrain} and Table \ref{tab:OtherTasks_finetune} show the results of adding code data in the pre-training phase and adding code instructions in the instruction tuning phase (only in this phase). Experimental results show that, in most cases, adding code data at both stages has little negative impact on the performance of other tasks and even produces some benefits. In the DuReader reading comprehension task, part of the performance will be reduced after adding code at different stages. This may be because the model does not thoroughly learn the code and text data, resulting in confusion when the model generates answers to reading comprehension questions. In future, we will verify and solve it in a larger model and with larger data.

\section{Related Work}

{\bfseries LLM training.} LLMs are usually based on the transformer architecture~\citep{dai2019transformer}. Notable models include BERT~\citep{devlin2018bert}, GPT-2~\citep{radford2019language}, and T5~\citep{raffel2020exploring}; after the emergence of GPT-3~\citep{brown2020language} with 175B parameters, a batch of larger models emerged, including PaLM~\citep{chowdhery2022palm}, OPT~\citep{zhang2022opt}, PanGu-$\alpha$~\citep{zeng2021pangu}, and LLaMA~\citep{touvron2023llama}, which have achieved remarkable results on various NLP tasks. For LLMs to follow instruction output, instruction tuning~\citep{peng2023instruction} plays an important role. This can use human-annotated feedback~\citep{ouyang2022training} or public benchmarks to automatically generate instructions~\citep{wang2022benchmarking, pCLUE} to fine-tune models on various tasks. 


{\bfseries Data Mixtures.} Models such as GPT-3~\citep{brown2020language} and PanGu-$\alpha$~\citep{zeng2021pangu} are trained on natural language data from various domains, and models such as LaMDA~\citep{thoppilan2022lamda} and LLaMA~\citep{touvron2023llama} are additionally trained on code data. However, the impact and specific origin of this mixed-code data is unclear. Some researchers have extensively analyzed the performance of current LLM on various tasks, pointing out that code may be the key to improving reasoning ability~\citep{liang2022holistic, fu2022gptroadmap}. However, the evaluated models have different parameters and data scales, and problems such as unknown training details exist. It is difficult to determine the exact impact of code data on the reasoning ability of LLMs.

\section{Conclusion}
In this paper, we investigate at which stage introducing code data can help improve the reasoning ability of LLMs. We validate the effect of code data at different stages with the same parameter scale and using the same training objective. We point out that simply adding code data in the pre-training phase can effectively improve the general reasoning ability of the model. Furthermore, we find that adding code instructions in the instruction tuning stage can make the model follow human instructions for output and improve specific code reasoning capabilities. Moreover, we point out that the dynamic mixing strategy of code and text data assists LLMs in learning reasoning capability step-by-step during the training process. We provide a well-designed and tested reference implementation for LLMs training to help researchers and developers better understand and analyze LLMs.



\bibliography{iclr2024_conference}
\bibliographystyle{iclr2024_conference}

\appendix
\section{Code Reproduction}
\subsection{Torch and Mindspore Version}
To ensure reproducibility, we open-sourced the model training and inference code using the Torch framework in the repository link:~\url{https://anonymous.4open.science/r/tmp-ADDE/README.md}. To facilitate reproducibility on the MindSpore platform and Ascend 910 hardware, we have deployed the code on the OpenI community platform (New Generation Artificial Intelligence Open Source Open Platform). This step aims to enable users to easily access and utilize MindSpore for training and inference. You can find the deployment at the following link: ~\url{https://openi.pcl.ac.cn/anonymous\_user/CodeLLMReasoning?lang=en-US}.

\subsection{Introduction to Mindspore}
We acknowledge that MindSpore is relatively newer and requires further development time. However, we would also like to highlight the adoption of MindSpore by prominent researchers in the field. For instance, ~\citep{zheng2023codegeex} in their work on CodeGeeX, \citep{liu2021opt}'s OPT model and \citep{christopoulou2022pangu} in the release of PanGuCoder have both embraced the software-hardware combination offered by MindSpore. According to the data from Papers with Code~(\url{https://paperswithcode.com/trends}), there have been 398 repositories utilizing the open-source MindSpore framework since 2023, which is higher than the 192 repositories using TensorFlow. This showcases its potential and the growing interest within the community.

\section{Experiments Details}
\label{sec:experimentsdetails}

\begin{figure}[t!]
    \centering
    \includegraphics[scale=0.40]{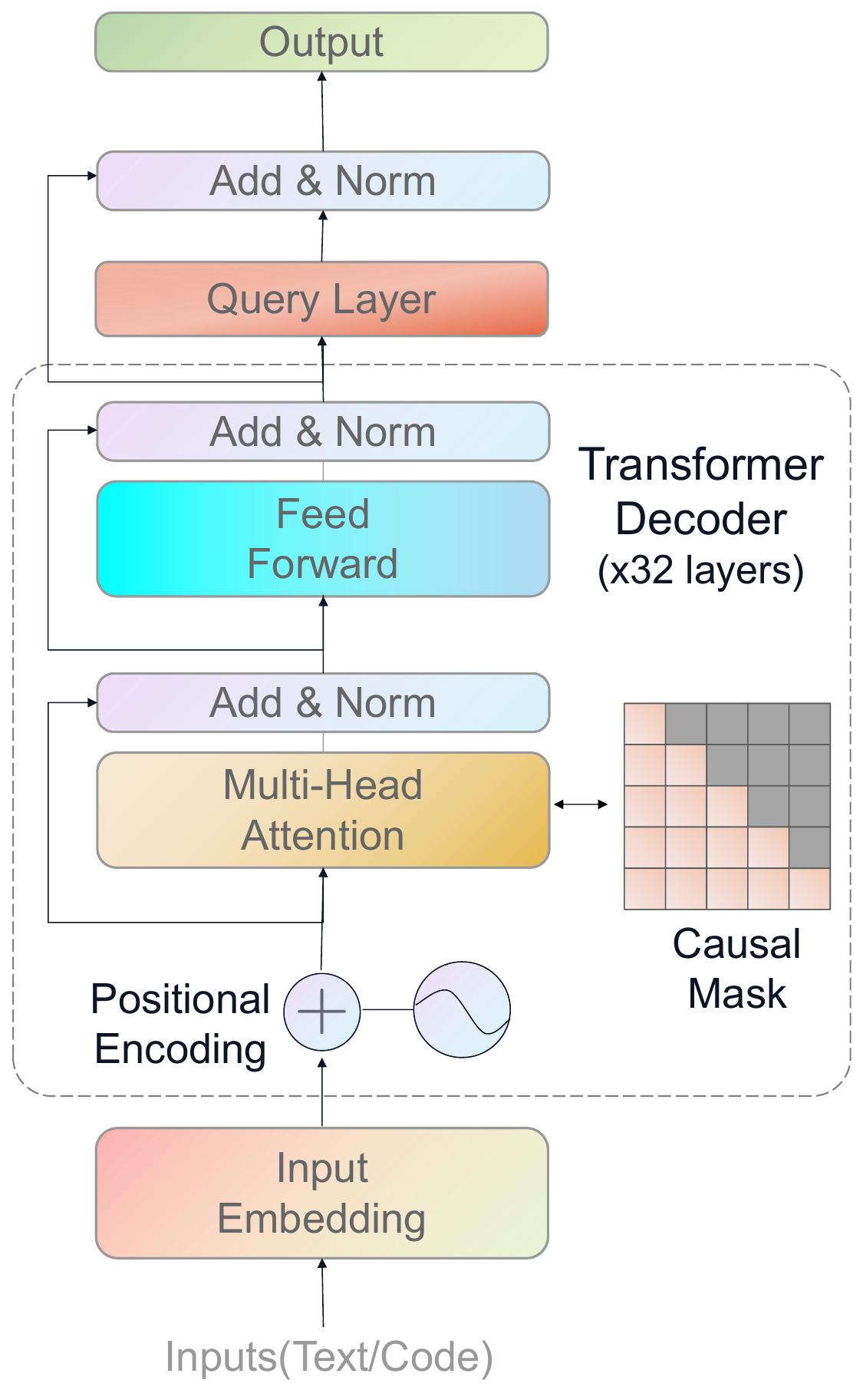}
    \caption{Model architecture. We build models with 2.6B and 13B parameters, consisting of 32-layer left-to-right transformer decoders and a top query layer.}
    \label{fig:fig2}
\end{figure}

Our experiments are developed under the Mindspore framework. To make a fair comparison with the baseline of PanGu2.6B, we retain the setting of the 32-layer transformer decoder. The model architecture as shown in Figure \ref{fig:fig2}. In the pre-training stage, we trained CodePanGu2.6B on a cluster of 16 Ascend 910 AI processors, and in the instruction-tuning stage, we tuned models on a cluster of 8 Ascend 910 AI processors. The sequence length for the training data is set to 1024 for all the models. Other detailed configurations can be found in Table \ref{tab:detail}. 

\begin{table}[h]
\centering
\renewcommand{\arraystretch}{1.5}
\begin{tabular}{ccc}
\hline
\textbf{Type} & \textbf{Parameter} &\textbf{Value}\\
\hline
\multirow{4}{*}{Environmental parameter} & Framework & Mindspore v1.7.0 \\
& Hardwares &  Ascend 910 \\
& Mem per GPU & 32GB \\
& GPUs per node & 8 \\ \hline
\multirow{4}{*}{Model parameter} & Layers &  32 \\
& Hidden size &  2560 \\
& FFN size &  10240 \\
& Heads &  32 \\\hline
\multirow{5}{*}{Optimization parameter} & Optimizer & Adam \\
& Initial/final learning rate &  1e-4(2e-5)/1e-6 \\
& Warm-up step & 500 \\
& Learning rate scheduler & cosine \\
& Optimizer parameters & $\beta1= 0.9, \beta2 = 0.95$ \\\hline
\multirow{3}{*}{Parallelism parameter}  & Data parallel & 16(8) \\
 & Model parallel & 1 \\
 & pipeline parallel & 1 \\\hline
\end{tabular}
\caption{Training configurations.(The values in parentheses are instruction-tuning parameters)}
\label{tab:detail}
\end{table}

\section{The Template for Other Tasks}
\label{sec:appendixA}

We follow Chapter \ref{evaluationdetails}, conduct experiments on other tasks to verify the impact of code data on other comprehension and generation tasks that are less demanding on reasoning, including $C^3$~\citep{sun2020investigating}; two NLI tasks (OCNLI~\citep{ocnli} and CMNLI~\citep{wang2018glue}); one reading comprehension task duReader~\citep{he2017dureader}. Table \ref{tab:promptother} shows the prompt templates for these tasks. The evaluation metrics for duReader, including F1 and exact match(EM), measure the similarity between the predicted and ground-truth text spans. The evaluation metric of other tasks is accuracy.

\begin{table}[h]
\centering
\renewcommand{\arraystretch}{1.5}
\scalebox{0.95}{
\begin{tabular}{cl}
\hline
\textbf{Task} &\textbf{Input \& Prompt}\\
\hline
$C^3$ & Question: \underline{\$question}\textbackslash n Answer:\underline{\$choice} comes from the dialogue: \underline{\$context} \\

OCNLI & \underline{\$S1}? Yes/Maybe/No, \underline{\$S2} \\
CMNLI &  \underline{\$S1}? Yes/Maybe/No, \underline{\$S2}  \\
duReader &  Read document: \underline{\$Document}\textbackslash n Question:\underline{\$Question} \textbackslash n Answer:  \\\hline
\end{tabular}}
\caption{The input \& prompt template for other tasks.}
\label{tab:promptother}
\end{table}

\section{Training Loss}
\begin{figure}[t]
    \centering
    \includegraphics[scale=0.8]{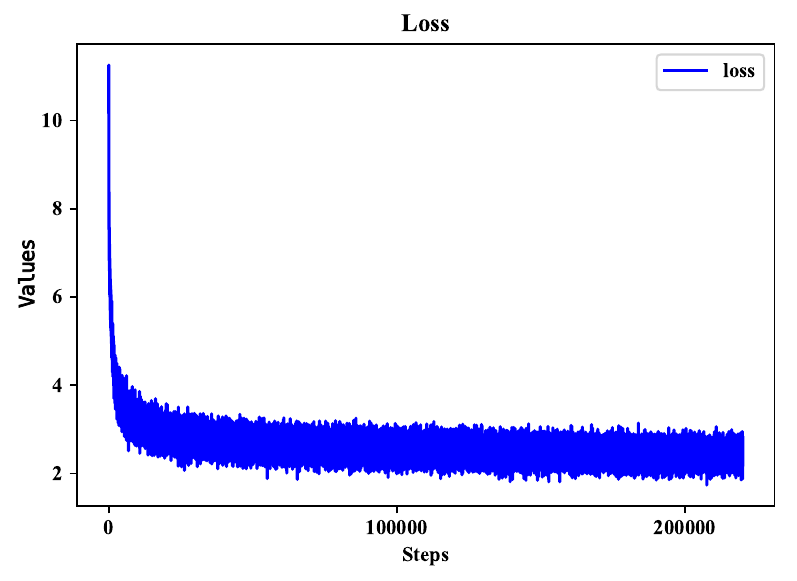}
    \caption{The curves of training loss for CodePanGu2.6B.}
    \label{fig:traincurve}
\end{figure}

The curves of training loss for the CodePanGu2.6B model are shown in Figure \ref{fig:traincurve}. We show that the cross entropy loss decreases steadily during training and the loss of this model converges to around 2.25.

\label{sec:trainloss}

\section{Case Study}  \label{casestudy}
\begin{figure*}[t!]
    \centering
    \includegraphics[scale=0.50]{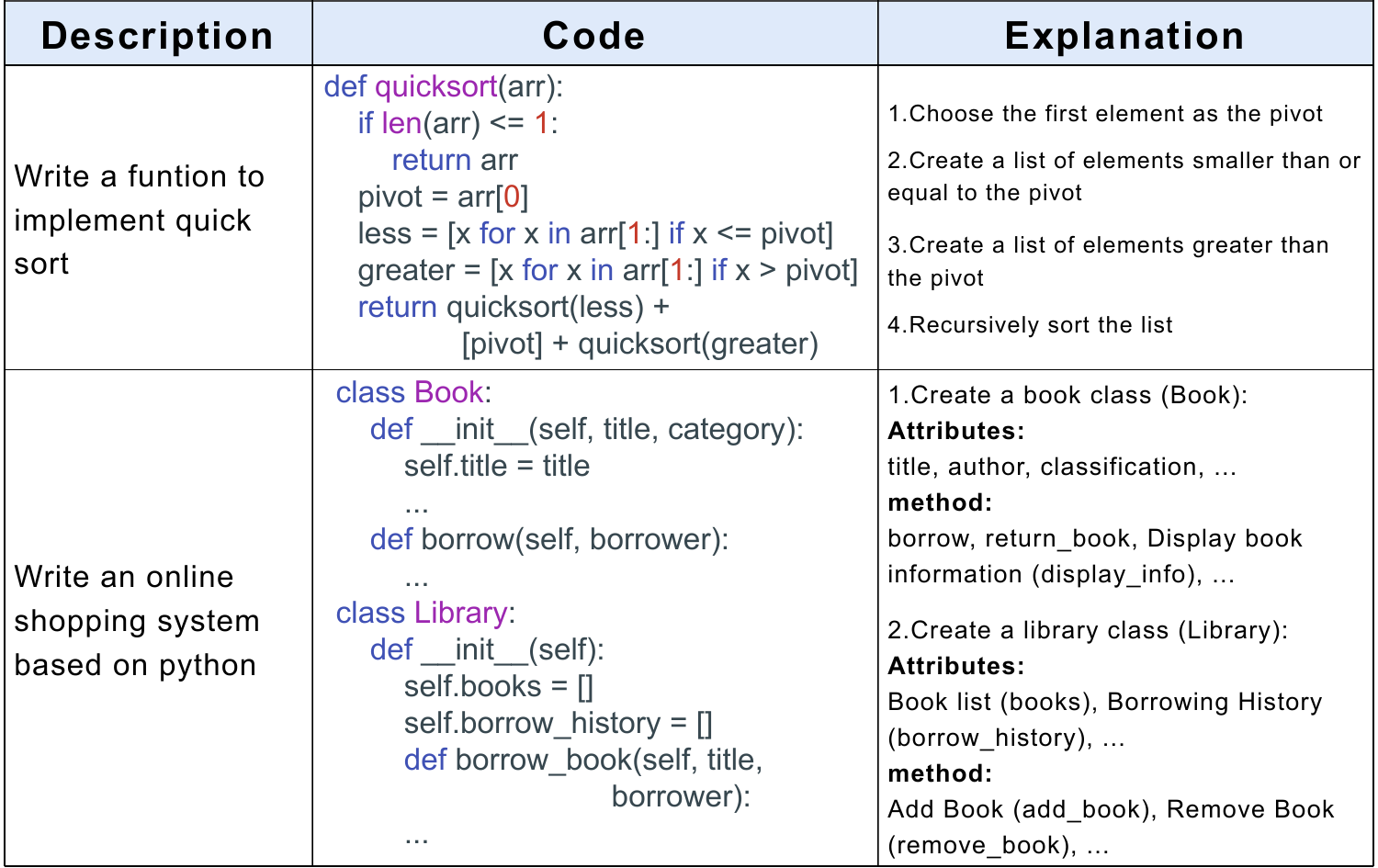}
    \caption{Examples of different codes.}
    \label{fig:case1}
\end{figure*}

In summary, adding code data in the pre-training stage can effectively improve the general reasoning ability of LLM, and can guide the model to make full use of the coherent reasoning process of the Chain-of-Thought to generate answers. Consistent with GPTRoadMap's point of view~\citep{fu2022gptroadmap}, we think this may have something to do with the logic of the code itself. To further explain why the code improves the reasoning ability of the model, we found several sample codes from the dataset and explained each code, as shown in Figure \ref{fig:case1}.

\begin{figure}[h]
    \centering
    \includegraphics[scale=0.70]{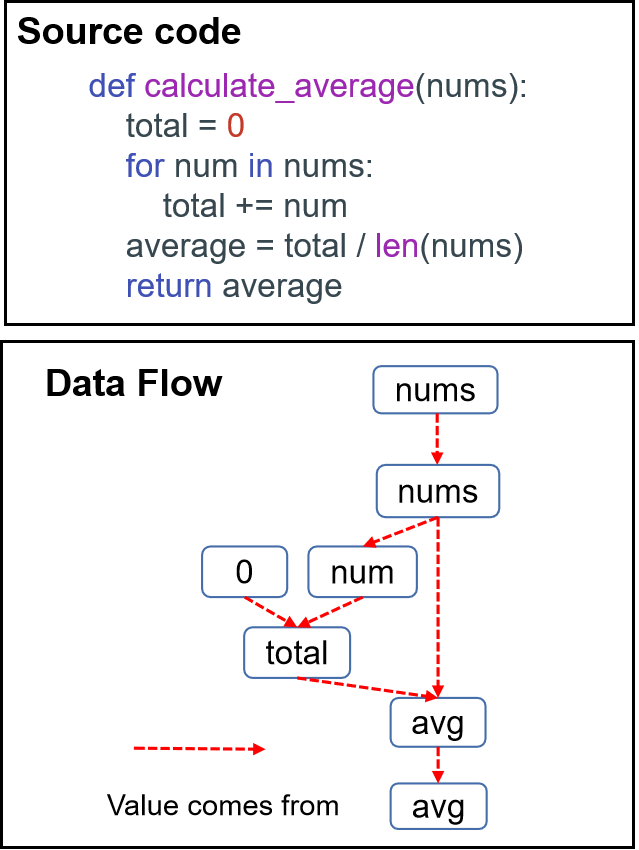}
    \caption{Examples of code dependencies.}
    \label{fig:case2}
\end{figure}

We found that, regardless of the length of the code dealing with different problems, step-by-step reasoning is required to ensure that the code is generated correctly, similar to the Chain-of-Thought required by other reasoning tasks. This may indicate that the model implicitly learns the thinking chain ability through the code data, which improves the reasoning ability of the language model. In addition, we analyzed the data flow graph of the \textit{calculate\_average} function, as shown in Figure \ref{fig:case2}. We found many data flow dependence relations in the code data, which are distributed among different code variables. Complex reasoning tasks usually require long dependencies to infer correct conclusions, so the language model may benefit from dependencies such as data and control flow of code data and improve the reasoning ability of the model.

\section{Dataset Construction}  \label{pangudata}
{\bfseries Cleaning and Filtering.} To improve the data quality, we adopt the following rule-based text cleaning strategies over the raw web pages from Common Crawl. 
\begin{itemize}

    \item Remove the document which contains less than 60\% Chinese characters.

    \item Remove the document which contains less than 150 characters.

    \item Remove the document which contains only the title of a webpage.

    \item Remove the special symbols and duplicated paragraphs in each document.

    \item Identify advertisements based on keywords and remove documents that contain advertisements.

    \item Identify the navigation bar of the web page and remove it.     

\end{itemize}

Regarding the use of The Common Crawl corpus~\url{https://commoncrawl.org/the-data/} in our work, it's worth noting that several prominent projects, such as Llama, have also employed this dataset. Crawled data may exhibit biases, and ground truth data often resides within large corporations, making access challenging. In light of this, we adopted a strategy similar to other open-source models like Llama and Falcon, leveraging a broader range of data types such as open-source code and e-books to supplement and mitigate potential biases.

{\bfseries Text Deduplication.} Although we removed duplicate paragraphs in each document in the previous step, there are still documents with highly overlapping content in different data sources. Therefore, we carry out fuzzy data deduplication over the documents across all our data sources to further remove high-overlap content. For fuzzy data deduplication, we employed  Spark's MinHashLSH algorithm, a widely adopted technique by models like GPT-3. We will revise our text to expound on these specifics.

{\bfseries Data Selection.} Using the construction process described above, we constructed filtered text corpora from five types of data sources. Based on this corpus, we constructed a training dataset of 100GB text data by sampling each data source according to the ratio of Figure \ref{contrastive} and used this data as the first part of the training set to train CodePanGu2.6B.

\section{URLs of Used Datasets}
This section gives the URLs of the used benchmark datasets. 

\begin{itemize}

    \item The Common Crawl corpus: https://commoncrawl.org/the-data/
    \item BAAI-WuDao: https://openi.pcl.ac.cn/BAAI/WuDao-Data
    \item CodeParrot: https://huggingface.co/codeparrot/codeparrot
    \item github-code: https://huggingface.co/datasets/codeparrot/github-code
    \item stanford\_alpaca: https://github.com/tatsu-lab/stanford\_alpaca
    \item code\_alpaca: https://github.com/sahil280114/codealpaca
    \item PromptCLUE: https://github.com/clue-ai/PromptCLUE
    

    
    
    

\end{itemize}

\section{Limitation}
\subsection{Verified on more large language models.}
PanGu~\citep{zeng2021pangu}, along with other LLMs like Llama~\citep{touvron2023llama} and Google PaLM~\citep{chowdhery2022palm}, shares a common architecture based on the GPT-2~\cite{chatgpt} decoder-only architecture and next token prediction task. Due to resource constraints and environmental concerns, we currently only conduct experiments on PanGu, but we compare models with different parameter scales (\ie 2.6B and 13B) to demonstrate the impact of the code. At the same time, we look forward to the emergence of LLMs with different architectures than GPT, and we also look forward to following up and verifying more LLMs with different architectures.

\subsection{Why not design a new model for code?}

Our paper primarily investigates at which training stage can codes help general LLMs reasoning. Through extensive experimentation, we assess the impact of code data, offering insights that can guide the development of future universal LLMs. Our objective is not to create a specialized code or text model, but rather to provide guidance in this context. In the future, we plan to leverage these findings for future development. We will explore integrating code features with models to create more robust and versatile universal reasoning models.

\end{document}